\documentclass{article}

     \usepackage[preprint]{neurips_2020}

\usepackage[utf8]{inputenc} 
\usepackage[T1]{fontenc}    
\usepackage{hyperref}       
\usepackage{url}            
\usepackage{booktabs}       
\usepackage{amsfonts}       
\usepackage{nicefrac}       
\usepackage{microtype}      

\usepackage{graphicx}
\usepackage{mathtools}
\usepackage{commath}
\usepackage{amsmath}
\usepackage{nccmath}
\usepackage[dvipsnames]{xcolor}
\usepackage{soul}
\usepackage{wrapfig,lipsum,booktabs}
\usepackage{amsfonts}

\title{Egocentric Object Manipulation Graphs}

\author{%
  Eadom Dessalene \\
  University of Maryland, College Park \\
  \texttt{edessale@cs.umd.edu} \\
   \And
   Michael Maynord \\
   University of Maryland, College Park \\
   \texttt{maynord@umd.edu} \\
   \AND
   Chinmaya Devaraj \\
   University of Maryland, College Park \\
   \texttt{chinmayd@cs.umd.edu} \\
   \And
   Cornelia Ferm\"{u}ller \\
   University of Maryland, College Park \\
   \texttt{fer@umiacs.umd.edu} \\
   \And
   Yiannis Aloimonos \\
   University of Maryland, College Park \\
   \texttt{yiannis@cs.umd.edu} \\
}

\begin{document}

\maketitle

\begin{abstract}

We introduce Egocentric Object Manipulation Graphs (Ego-OMG) - a novel representation for activity modeling and anticipation of near future actions integrating three components:  1) \textit{semantic temporal structure} of activities, 2) \textit{short-term dynamics}, and 3) representations for \textit{appearance}. \textit{Semantic temporal structure} is modeled through a graph, embedded through a Graph Convolutional Network, whose states model characteristics of and relations between hands and objects. These state representations derive from all three levels of abstraction, and span segments delimited by the making and breaking of hand-object contact. \textit{Short-term dynamics} are modeled in two ways: A) through 3D convolutions, and B) through anticipating the spatiotemporal end points of hand trajectories, where hands come into contact with objects. \textit{Appearance} is modeled through deep spatiotemporal features produced through existing methods. We note that in Ego-OMG it is simple to swap these appearance features, and thus Ego-OMG is complementary to most existing action anticipation methods. We evaluate Ego-OMG on the EPIC Kitchens Action Anticipation Challenge. The consistency of the egocentric perspective of EPIC Kitchens allows for the utilization of the hand-centric cues upon which Ego-OMG relies. We demonstrate state-of-the-art performance, outranking all other previous published methods by large margins and ranking first on the unseen test set and second on the seen test set of the EPIC Kitchens Action Anticipation Challenge. We attribute the success of Ego-OMG to the modeling of semantic structure captured over long timespans. We evaluate the design choices made through several ablation studies. Code will be released upon acceptance.

\end{abstract}

\section{Introduction}

	Computer vision as a field has advanced rapidly in the past decade. The focus of this advancement has centered primarily around understanding visual structure in static images. Recently, developments in hardware enabling faster video processing and an increasing saturation of work on static images has led to an increased interest in video. By-and-large this interest has produced work which attempts to translate paradigms developed for image understanding onto video understanding. We seek to extend beyond such work through a rich modeling of temporal semantic structure in video.
	
	Human agents are the primary drivers of meaningful change in video - as such we center our work around human action. The defining characteristics of an action are often semantic and relational rather than appearance based in nature – as such, methods which rely solely on appearance modeling are not well adapted to representing action.
	

	Most work on action understanding has centered on the task of action recognition: shown a clip of an action, can the class of that action be inferred? We instead work on the task of action anticipation: the task of classifying future actions from observations preceding the start of the action. We focus on human manipulation activities in video. This is arguably the most compelling – and the most challenging – of video domains to model. It is compelling as it leads towards research on human centered applications - agents which understand and / or interact with ourselves. It is challenging as human behavior is more challenging to model than the dynamics of inanimate objects. We employ ego-centric video for the advantages it provides: a consistent visual perspective of human actions; and, a perspective in which the hands engaging in manipulation are more clearly visible.

	There are both theoretical and practical motivations to the task of action anticipation. On a theoretical level, action anticipation requires a modeling not only of the appearance characteristics of video but the dynamic characteristics as well – characteristics which are absent from static images. As such, the task of action anticipation, as opposed to recognition, provides sound benchmarks with which to evaluate a method's ability to model temporal dynamics and semantics.

    As for practical motivations, action anticipation is of central importance for surveillance, navigation and human-robot interaction systems. For example, robots being able to provide earlier feedback based on accurate predictions of what humans are about to perform can help reduce both the physical and cognitive load on humans performing the task. Additionally, for robots to preempt human actions and respond accordingly produces a more natural, human-like interaction. 

	We select EPIC Kitchens as our dataset, both because of the suitability of cooking activities to action anticipation, and because of the existing EPIC Kitchens Action Anticipation Challenge.

	Appearance is an important, though not sufficient, element in understanding video – we leverage existing work in appearance based video understanding. Namely, we employ CSN \cite{Tran_2019_ICCV} features - a variant of I3D \cite{carreira2017quo} features that make use of channel-wise group 3D convolutions - capturing not only static visual appearance but short time-frame motion characteristics.


	To extend further beyond the prediction of the short term world dynamics capturable in the time-frame of CSN features, we leverage a characteristic of the EPIC Kitchens dataset: it consists of ego-centric video in which the meaningful action elements consist of hands interacting with objects. To extend beyond CSNs we anticipate contact events through the modeling of short term hand dynamics, producing soft segmentation masks corresponding to the hands, objects in contact, and anticipated objects of interaction.
	
	


We model the progression of these interactions with states delimited by contact events, capturing characteristics of and relations between hands and objects. We represent states as the composite interaction of two hands with the possible objects in the environment. Connecting state to state through transitions produces a graph, which is embedded into a Graph Convolutional Network (GCN) to produce natural vector state representations. For purposes of extended modeling of relations, we feed this state history into an LSTM, on top of which action anticipation is performed.

We achieve \textbf{\textit{1st}} place on the unseen test set of the EPIC Kitchens Challenge, reaching major improvements over previous state of the art methods \cite{furnari2019would,liu2019forecasting}. We achieve \textbf{\textit{2nd}} place on the seen test set, outperformed only by the anonymous \textbf{action\_banks} submission, outperforming all previous published submissions. We find in addition leveraging graph embeddings for action anticipation alone is enough to outperform end-to-end video architectures at anticipating actions over long time horizons.


    The primary contributions of Ego-OMG are:
    \begin{itemize}
        \item A natural integration of representations at multiple levels of abstraction: appearance, dynamic, semantic.
        \item A novel graph based representation for manipulation activity, whose states derive from appearance, dynamic, and relations characteristics of hands and objects.
        \item Complementarity with existing methods through a modular appearance and short-term dynamics based stream.
        \item A surpassing of previous state-of-the-art on the EPIC Kitchens Action Anticipation challenge, achieving 1st place on the Unseen test set and 2nd place on the Seen test.
    \end{itemize}


	The remainder of this paper is structured as follows: In Section \ref{sec:related_work} we cover related works. In Section \ref{sec:structured_graph_of_egocentric_activity} we introduce our proposed graph representation, and in Section \ref{sec:joint_action_anticipation_architecture} propose our joint action anticipation architecture. In Section \ref{sec:experiments} we discuss the experiment setting and go over ablations evaluating system components. Finally, in Section \ref{sec:conclusion} we conclude.

\begin{figure}
\includegraphics[width=1\textwidth]{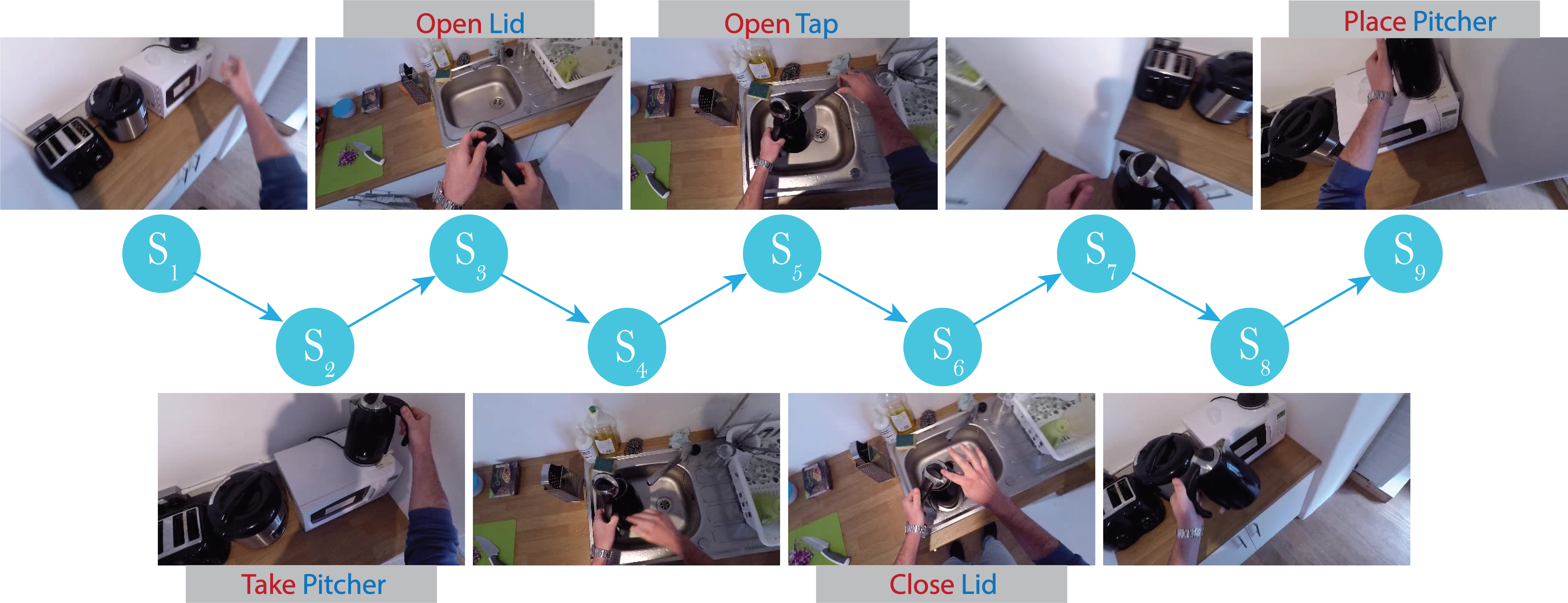}
  \centering
  \caption{The activity "refill pitcher" is shown above, with associated action labels shown. Despite the activity lasting several minutes and containing multiple action segments, Ego-OMG structures the video clip into a discrete set of states $S_k = \{s_1, s_2, ..., s_n\}$, where each state represents the objects presently in contact or anticipated to soon be in contact with the left and right hands.
  }
\end{figure}


\section{Related Works}
\label{sec:related_work}

\paragraph{Action Anticipation}
Action anticipation is the task of classifying future actions from observations that end before the actions begin. Action anticipation in third person video has been the focus of most previous works, typically hallucinating visual representations of future frames \cite{miech2019leveraging, zeng2017visual, gammulle2019predicting, vondrick2016anticipating} or modelling future dynamics of structured representations such as skeletal pose \cite{tanke2019human}, pedestrian trajectories \cite{xu2018encoding}, recipes \cite{sener2019zero}, etc. The release of large egocentric video datasets \cite{damen2018scaling, li2018eye} has lead to a re-focusing towards incorporating egocentric cues~\cite{liu2019forecasting, furnari2017next, nagarajan2020ego} and augmented learning architectures \cite{furnari2019would,liu2019forecasting} for handling the additional challenges that arise in the egocentric setting. 

\paragraph{Video Representation}
Typical works within action understanding involve two-stream architectures where the input to the network is RGB video fed to the network in parallel with pre-computed frames of optical flow \cite{carreira2017quo, simonyan2014very}. These approaches have achieved success in tasks where appearance and short-term motion is sufficient for the task at hand (i.e. action recognition) \cite{carreira2017quo}. However, it has been reported \cite{carreira2017quo, liu2019forecasting, li2018eye} that such methods do not 
transfer well to tasks such as action prediction or action anticipation. We find this understandable, as action anticipation requires reasoning about complex semantic cues that go beyond appearance.

Rather than simply represent the video as a stack of frames, it is desirable to capture the long-term semantics underlying the video observation of the activity. Recent works have proposed the enrichment of raw video features with graphs \cite{jain2016structural, Wang_2018_ECCV, nagarajan2020ego, soran2015generating}. Typically graph nodes represent detected objects, actors, or locations. Unlike other works that utilize an exhaustive list of entities, by restricting ourselves to the modelling of objects either currently or expected to be in contact with the hands, we are able to rule out 'background' objects that play no role in the actions involved, effectively using the hands as an attention mechanism.


\paragraph{Structured Learning}
Graphical models have been a classic solution to many problems in computer vision involving the modelling of entities and their respective relationships such as action prediction~\cite{lan2014hierarchical}, video summarization~\cite{izadinia2012recognizing}, articulated pose estimation~\cite{pfister2015flowing} and most commonly semantic segmentation~\cite{zheng2015conditional}. Recent years have seen the rise of Convolutional Neural Network (CNN) architectures for learning over graphs. Such approaches allow the semi-supervised learning of low-dimensional vector embeddings of nodes in large graphs, arriving at useful node representations for downstream inference tasks. In this work we apply Graph Convolutional Networks (GCNs)~\cite{kipf2016semi}, the generalization of CNNs to graphs of arbitrary structures. Previous works have applied GCNs to the task of action recognition~\cite{Wang_2018_ECCV, yan2018spatial, chen2019graph}, achieving state-of-the-art results on a wide variety of video datasets. We do the same in the action anticipation setting.


%

\begin{figure}
\label{fig:architecture}
\includegraphics[width=1\textwidth]{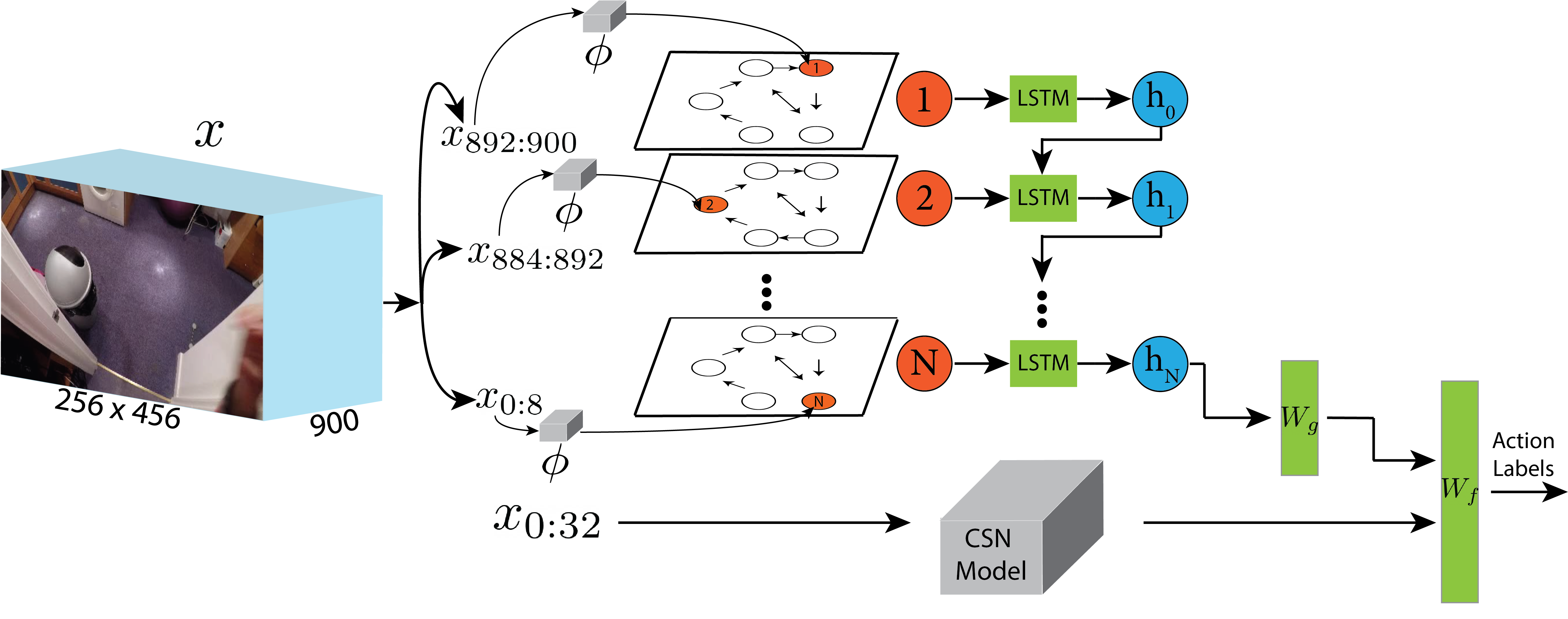}
  \centering
  \caption{Overview of Ego-OMG's architecture. Ego-OMG consists of two streams: 1) The top stream consists of the extraction of a discretized sequence of states from an unconstrained egocentric video clip $x$ of $900$ frames using the Contact Anticipation Network $\phi$. The nodes predicted by $\phi$ are embedded through GCN layers and then fed to an LSTM. This is then followed by a 1-layer MLP $W_g$ to generate softmax scores for the anticipated future action. 2) The second stream generates softmax scores for the anticipated future action through feeding a short history (the last 32 frames of $x$) of video to a CSN model. A 1-layer MLP $W_f$ processes the concatenated L2-normalized softmax scores to perform action anticipation.}
\end{figure}

\section{Joint Action Anticipation Architecture}
\label{sec:joint_action_anticipation_architecture}


The architecture of Ego-OMG is shown in Figure \ref{fig:architecture}. Input consists of a single clip spanning 60 seconds - or 900 frames. This clip spans from start time $\tau_s - 60$ seconds to end time $\tau_s$ seconds, where $\tau_s$ lies $\tau_a$ seconds before the start of the action to be anticipated. The output consists of a logit layer predicting the class of the action $\tau_a$ seconds after the end of the observation. The architecture is comprised of two streams: One modeling the appearance and short term dynamics of the last few seconds of the clip; the other modeling hand dynamics and long-term semantic temporal relations.




In the first stream, we model appearance and short-term dynamics with a Channel-Separated Convolutional Network (CSN), a 3D CNN factorizing 3D convolutions in channel and space-time in similar fashion to Xception-Net \cite{chollet2017xception} which factorizes 2D convolutions in channel and space. The weights are pre-trained on the largescale IG-65M video dataset \cite{ghadiyaram2019large}. The network takes as input 32 consecutive frames of size $T\times256\times256$, where $T = 15$ is the number of frames and $256$ is the height and width of the cropped inputs. We apply horizontal flipping, color jittering and random crops during training, with centered crops during testing. The model is trained using SGD with a batch size of 16, a learning rate of $2.5 \times 10^{-3}$ and a momentum of 0.9.

In the second stream we model dynamics of interactions between hands and objects, as well as longer term temporal semantic relations between the actions of the activity. We capture this structure in the form of a graph, described in detail in Section \ref{sec:structured_graph_of_egocentric_activity}. After computing the graph over the entire EPIC Kitchens dataset, we feed it through two graph convolution layers of hidden layer size $256$ and $128$ respectively. Note our application of the GCN is transductive; it is applied on a single, fixed graph consisting of all nodes seen during train \textit{and} test time beforehand. The GCN training achieves fast convergence, reaching peak validation accuracy after $5$ epochs or roughly $0.25$ hours of training on a NVIDIA GeForce GTX 1080 GPU. At test time, we convert an input video of $900$ frames to a sequence of states and from each state's respective node embedding $g_n$ for $n \in N$, we aggregate the state history with a 1-layer LSTM. From the final hidden state $h_N$, we apply a 1-layer MLP $W_g$ to classify the next most likely action. We feed the sequence of node embeddings into an LSTM \cite{hochreiter1997long}. The LSTM carries hidden states of size $128$. A batch size of 16 and a learning rate of $7 \times 10^{-5}$ is used with ADAM optimizer. 

We concatenate the L2-normalized softmax scores from each respective stream, freezing the two sub-networks while training a 1-layer MLP $W_f$ with a batch size of $16$ and learning rate of $0.01$ on top of the joint softmax scores to classify the next most likely action. We find a late fusion approach provides slight benefits in practice as opposed to an early fusion of the two streams, likely due to the different learning dynamics of the individual streams.

\begin{figure}
  \begin{center}
  \label{fig:graph}

\includegraphics[width=1\textwidth]{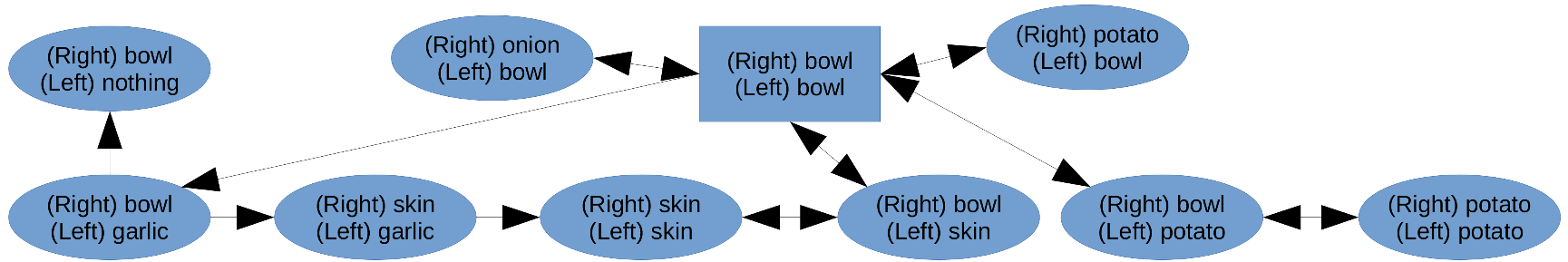}
  \end{center}
  \caption{Example graph derived from a 30 second EPIC Kitchens clip illustrating the semantic relations which Ego-OMG models. Note that Ego-OMG leverages longer video sequences of activity. For easier illustration we include in states only the object in direct contact with each hand (full states include objects involved in anticipated contact). The rectangular node denotes is the starting state.}
    \vspace{-10pt}
\end{figure}




\section{Structured Graph of Egocentric Activity}
\label{sec:structured_graph_of_egocentric_activity}

Action anticipation involves reasoning about complex contextual and temporal semantic relations. We design a graph based representation to capture these relations.


The structure of an egocentric activity video dataset is consolidated into graph $G$, from which we can retrieve a high level sequence of node states $\tau = x_{1:N}$ extracted from video $V_n$ at test time and perform soft reasoning upon to anticipate the next most likely action.

Given the egocentric setting and the general observation that hands are the central driving force of change in object manipulation activities, graph states represent \textit{contact} and \textit{anticipated contact} relations between hands and objects, where each hand is represented independently. This enables us to more finely model tasks requiring complex bi-manual manipulation, as illustrated in Figure \ref{fig:graph}.

\subsection{Contact Anticipation}
\label{subsec:contact_anticipation}

As illustrated in Figure \ref{dataset} part b), we feed $x_t$ into a \textit{Contact Anticipation Network} \cite{edessale2020cont} $\phi$, whose purpose is to anticipate hand object contacts. Anticipated contact is represented through a $4$ channel output, consisting of object segmentation masks $\{\Psi_{t_r}, \Psi_{t_l}, \Gamma_{t_r}, \Gamma_{t_l}\}$, where $\Psi_{t_r}$ and $\Psi_{t_l}$ denote the next active object predictions, and $\Gamma_{t_r}$ and $\Gamma_{t_l}$ denote the objects detected to be presently in contact with the hand, both for the right and left hands respectively. We feed each segmentation frame to an object classifier, arriving at predicted object classes $o_t = \{\psi_{t_r}, \psi_{t_l}, \gamma_{t_r}, \gamma_{t_l}\}$. We note that for the purposes of this work we predict up to $1$ object each for $\psi_{t_r}$, $\psi_{t_l}$, $\gamma_{t_r}$, and $\gamma_{t_l}$. This limitation prevents us from modelling scenarios where multiple objects are held by the same hand for tasks requiring dexterous manipulation (though, these are uncommon scenarios).

The contact anticipation network $\phi$ is a 3D ConvNet video object segmentation architecture that makes use of additional supervisory signals corresponding to time of progression of directed hand movements along with ground truth segmentation masks of objects of interaction. It takes as input $x_t$ with $8$ frames of video along with a self-generated history of contact anticipation maps, a fine-grained representation of \textit{time-to-contact} between the hands and each pixel belonging to the environment. \cite{edessale2020cont} further describes the details of how the contact anticipation module is trained.

In practice, while the contact anticipation module succeeds at localizing contacted objects, the classifier tends to mis-classify currently held objects due to the severe occlusion imposed by the hand, especially for small objects like scissors and utensils. Therefore, in building the graph we impose the constraint that every object currently contacted by each hand \textit{must have been anticipated} at some previous instance in time, before the presence of occlusion. In classifying the objects currently in contact with the hand, we take the intersection of top-5 object class predictions for that object with the object classes previously predicted in anticipation within the past $7$ seconds.

\subsection{Graph Construction}
\begin{figure}
\includegraphics[width=1\textwidth]{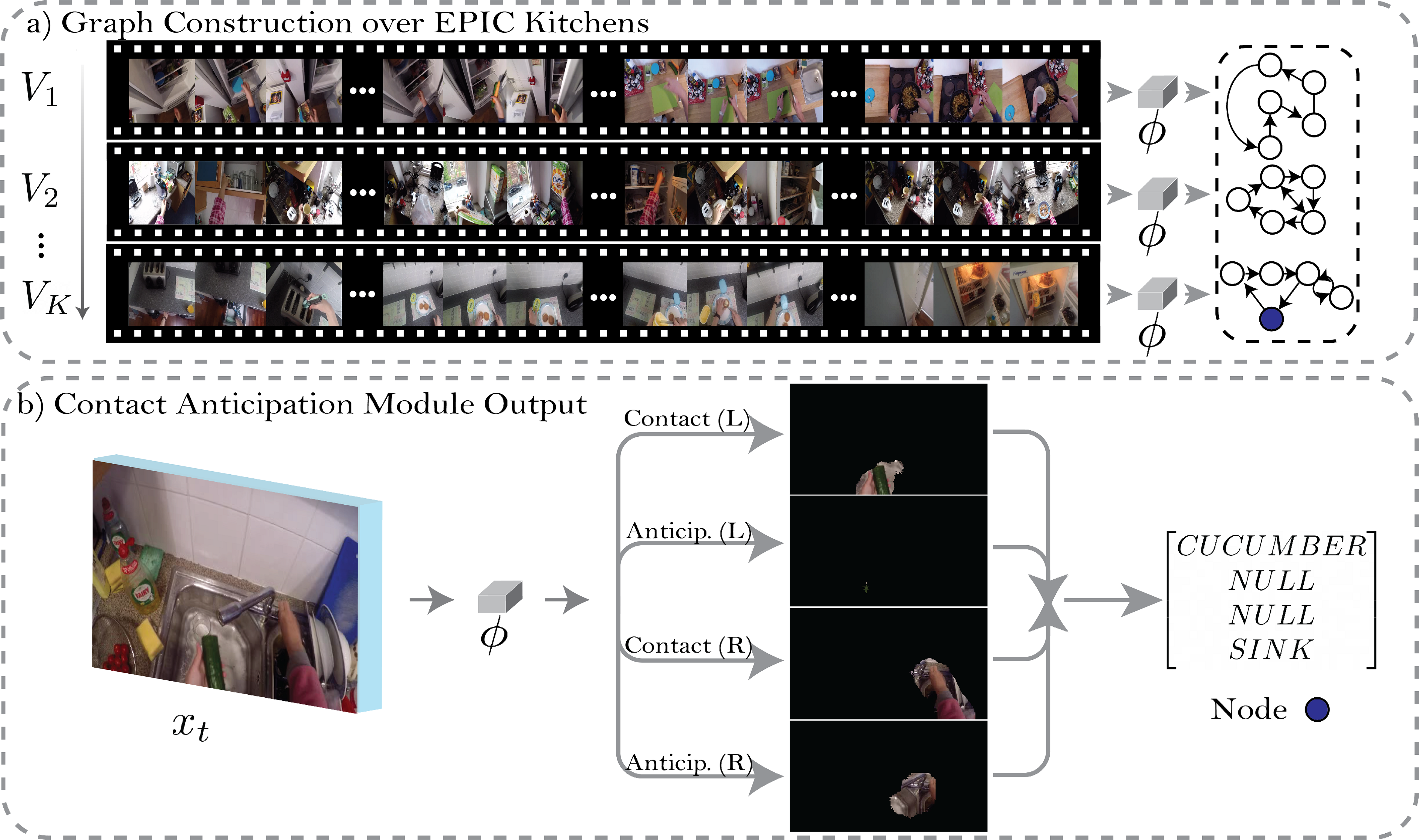}\label{dataset}
  \centering
  \caption{Overview illustrating the construction of a consolidated graph $G$ of hand-object interactions over a set of $K$ egocentric unconstrained videos. In (a), the Contact Anticipation Network $\phi$ is applied in a sliding window fashion over each video $V_k$, whose outputs are then classified w.r.t. object classes, and then consolidated into a single graph. (b) illustrates $\phi$ applied to a single window $x_t$ of $8$ frames where the output of the anticipation network is shown as $4$ separate binary segmentation masks representing the pixels belonging to the objects in contact with the left and right hand, and the pixels belonging to the objects anticipated to come into contact with the left and right hand. Each segmentation channel is fed to a classifier, and the resulting tuple is represented as a node in the graph.}
  \vspace{-10pt}
\end{figure}

As illustrated in Figure \ref{dataset} part a), we have a set of $K$ training videos, $V = \{V_1, V_2, ..., V_K\}$. To detect the objects involved in interaction, which are needed to build the graph, we utilize the Contact Anticipation Network, $\phi$, described in subsection \ref{subsec:contact_anticipation}. The network $\phi$ iterates over each video $V_i \in V$ using a sliding window with an 8-frame width, and a stride of 2. Feeding each of $4$ output channels of $\phi$ to the object classifier then produces detections $O_i = \{o_1, o_2, ..., o_{T_i/2}\}$ for $V_i$, where $T_i$ is the frame count of $V_i$. From the per-frame predictions of the object classes $o_i$, we suppress consecutive duplicate predictions, where $S_k = \{o_j: j=T_i/2-1 \lor o_j \neq o_{j+1}\}$ for $0 \leq j < T_i/2$, arriving at non-repeating states $S_k = \{s_1, s_2, ..., s_n\}$, an ordered set where temporal order is preserved.

With the input to graph construction defined, we now consider the graph $G = (V, E)$, where $E$ consists of the set of all edges, and $V$ consists of the set of all nodes. $V = \{V_s, V_a\}$ consists of nodes of two types: state nodes, and action nodes. State nodes consist of the intersection of all $S_k$, that is: $V_s=\bigcap_{k=1}^{K} S_k$, and action nodes $V_a$ consist of the set of all action classes $a_i \in A$, where $A$ is the set of all actions. In doing so, we represent both states and actions in graph $G$.

We construct the adjacency matrix as follows. Each node has an edge connecting it to itself: $e_{ii} \in E$ for $1 \leq i \leq |V|$ with weight $1$. We add weighted directed edges $e_{ij} \in E$ for consecutive states $s_i$ and $s_j$ for $0 \leq i < n$ and $j = i + 1$, where the weight $\sigma_{ij}$ is transition probability $p(s_{i+1} | s_i)$. We also add weighted directed edges between states and actions by adding weighted edge $e_{ij} \in E$ if action $i$ takes place within the timespan of state $s_j$, where weight $\sigma_{ij}$ is equal to $p(a_i | s_j)$.

Graph $G$ has a total number of nodes equal to the number of unique states $z = \abs{S} + \abs{A}$, where $S$ is the set of unique states and $A$ is the set of annotated actions. Let $X \in R^{z \times m}$ be a matrix containing all $z$ nodes with their corresponding features of dimension $m$. Rather than set $X$ to identity matrix $I$, we initialize each node with feature embeddings extracted from a pre-trained GloVe-600 model \cite{pennington-etal-2014-glove}. When representing states $s \in S$, we average the feature embeddings from each object noun in $s$. When representing actions $a \in A$, we average the embeddings for the verb and noun embeddings. We find that utilizing pretrained word embeddings for $G$ results in substantial performance gains over using $X = I$.

We feed the weighted adjacency matrix and $X$ as input into the GCN as described in Section \ref{sec:joint_action_anticipation_architecture}.

\begin{wraptable}{r}{6.5cm}
\vspace{-30pt}

   \caption{Action anticipation results on the EPIC Kitchens test set for \textit{seen} kitchens (\textbf{S1}) and \textit{unseen} kitchens (\textbf{S2}) during the EPIC Kitchens Action Anticipation Challenge. Only published submissions are shown.}
   \vspace{5pt}
    \label{table:comparison_other_methods}
   \centering
        \begin{tabular}{ r|c|c|c }
        \multicolumn{1}{r}{}
         &  \multicolumn{1}{c|}{Method}
         & \multicolumn{1}{c}{Top-1}
         & \multicolumn{1}{c}{Top-5}\\
        \hline{}
         & 2SCNN (RGB) \cite{shou2016temporal} & 4.32 & 15.21\\
         & TSN (RGB) \cite{wang2016temporal} & 6.00 & 18.21\\
         & TSN + MCE \cite{furnari2018leveraging} & 10.76 & 25.27 \\
        S1 & RULSTM\cite{furnari2019would} & 15.35 & 35.13 \\
         & Camp. et al\cite{camporese2020knowledge} & 15.67 & \textbf{36.31} \\
         & Liu et al\cite{liu2019forecasting} & 15.42 & 34.29 \\
         & \textbf{Ego-OMG} & \textbf{16.02} & 34.53 \\
        \hline{}
         & 2SCNN (RGB) \cite{shou2016temporal} & 2.39 & 9.35 \\
         & TSN (RGB) \cite{wang2016temporal} & 2.39 & 9.63 \\
         & TSN + MCE \cite{furnari2018leveraging} & 5.57 & 15.57 \\
        S2 & RULSTM\cite{furnari2019would} & 9.12 & 21.88 \\
        & Camp. et al\cite{camporese2020knowledge} & 9.32 & 23.28 \\
         & Liu et al\cite{liu2019forecasting} & 9.94 & 23.69 \\
         & \textbf{Ego-OMG} & \textbf{11.80} & \textbf{23.76} \\
        \end{tabular}

 \vspace{-30pt}

\end{wraptable} 

\section{Experiments}
\label{sec:experiments}

We evaluate Ego-OMG over the EPIC Kitchens Action Anticipation Challenge. In this section we report experimental details and results.


\subsection{EPIC Kitchens Dataset}
The EPIC Kitchens dataset is a large egocentric video dataset, captured by 32 subjects in 32 different kitchens. The videos consist of daily kitchen activities where participants were simply asked to record their interactions in their native kitchen environments (no instructional scripts were given to the subjects). Each video contains several annotated action segments, each associated a $(verb, noun)$ action label, where there are 125 unique verbs and 352 unique nouns, making for a total of 2,513 unique actions. The task is as follows: Given the boundaries of an action segment $[\tau_s - (\delta_o + \tau_a), \tau_s - \tau_a]$, we predict the anticipated action class $a_i$ by observing the video segment of observation length $\delta_o$ preceding the action start time $\tau_s$ by anticipation time $\tau_a$.

For the experiments comparing our approach with other state of the art action anticipation methods, we evaluate our results on the EPIC Kitchens test videos, consisting of two set sets: \textit{seen} kitchens (S1) and \textit{unseen} kitchens (S2). The seen kitchens split reflects videos where each kitchen is seen at train and test time, and the unseen kitchens split reflects videos where all sequences involving the same kitchen are either in the train split or test split. For the experimental analysis of our approach, we use the training videos to create our validation set, randomly splitting the public EPIC-Kitchens training set into 232 videos for training and 40 videos for validation as in \cite{furnari2019would}.

\begin{wraptable}{r}{5cm}
\vspace{-30pt}

  \caption{Action anticipation results over validation set averaged over $3$ runs for CSN stream, GCN stream, and CSN + GCN stream for \textit{seen} kitchens (\textbf{S1}) and \textit{unseen} kitchens (\textbf{S2}).}
    \centering
         \begin{tabular}{ r|c|c }
        \multicolumn{1}{r}{}
         &  \multicolumn{1}{c|}{Method}
         & \multicolumn{1}{c}{Accuracy} \\
        \hline{}
         & CSN & 15.51 \\
        S1 & GCN &  12.94\\
         & CSN + GCN &  \textbf{19.28}\\
        \hline{}
         & CSN &  9.82\\
        S2 & GCN &  10.10\\
         & CSN + GCN &  \textbf{13.06}\\
        \end{tabular}
  \label{table:gcn_csn_s1_s2}
\end{wraptable}

\subsection{Comparison to the State-of-art}
The protocol behind the EPIC Kitchens Action Anticipation Challenge is to set the action anticipation time $\tau_a$ to $1$ second. While there are 44 participants in the challenge, we report the top $3$ published submissions and include the original benchmarked action anticipation results from the EPIC Kitchens dataset release. We note that each alternative submission is \textit{complementary} to our submission; One could ideally replace our CSN model with any of the alternative top performing video architectures and expect to achieve even greater performance.

Table \ref{table:comparison_other_methods} illustrates performance comparisons between Ego-OMG and competing approaches. We observe Ego-OMG outperforms all other methods except for Top-5 action anticipation prediction for the seen test set (S1) - we conjecture that this is because \cite{furnari2018leveraging} and \cite{camporese2020knowledge} explicitly model ambiguity in their loss function, whereas we employ a standard categorical cross-entropy loss function.

\begin{table}[!t]
\caption{Action anticipation results over validation set for CSN stream, GCN stream and CSN + GCN stream over varying anticipation times $\tau_a$ seconds before the start of the action.}\label{table:anticipation_time_vary}

\centering
\begin{tabular}{|l*{6}{c}r|}
\hline
 \multicolumn{7}{ |c| }{Top-1 Accuracy over varying $\tau_a$ seconds anticipation time} \\
\hline
\multicolumn{1}{|r}{} & 5 & 2.5 & 1.5 & 1 & 0.5 & \multicolumn{1}{r|}{0}\\
\hline
CSN & 6.49 & 11.39 & 14.09 & 15.50 & 18.61 & \multicolumn{1}{r|}{19.37}  \\
GCN & 9.05 & 10.47 & 11.31 & 12.81 & 13.76  & \multicolumn{1}{r|}{14.56}  \\
CSN + GCN & \textbf{9.44} & \textbf{15.01} & \textbf{17.02} & \textbf{19.20} & \textbf{20.29} &  \multicolumn{1}{r|}{\textbf{21.89}}  \\
\hline
\end{tabular}
\vspace{-10pt}

\end{table}

\subsection{GCN Ablation}

Here we describe an ablation study evaluating the following questions pertaining to design choices made in constructing Ego-OMG:

\begin{enumerate}
  \setlength{\itemsep}{0pt}
  \setlength{\parskip}{-1pt}

\item What utility does each stream of Ego-OMG provide?
\item What utility does the LSTM provide in producing useful representations of state history for action anticipation?
\item What utility do graph embeddings effected through GCN layers applied over sequences $S_k = \{s_1, s_2, ..., s_n\}$ provide for action anticipation?
\item What utility do word embedding provide when representing world states $s_i \in S_k$?  
\end{enumerate}

\begin{wraptable}{r}{5cm}

\caption{Action anticipation accuracies from ablation studies over validation set with anticipation time $\tau_a = 1$ second, over different representations of state history.}\label{wrap-tab:1}
\begin{tabular}{l*{1}{c}r}\\
Method & Top-1 Acc.  \\
\hline
LSTM-Aggr. & \textbf{12.81}  \\
Term-State-Class.       & 11.70  \\
Mean-Aggr.            & 9.74 \\
\end{tabular}

\vspace{20pt}

\caption{Action anticipation accuracies over validation set with anticipation time $\tau_a = 1$ second, over GCN and GloVe embedding ablations.}
  \vspace*{.5\baselineskip}
 \begin{tabular}{ r|c|c }
\multicolumn{1}{r}{}
 &  \multicolumn{1}{c|}{GCN}
 & \multicolumn{1}{c}{No GCN} \\
\hline{}
GloVe Vectors & \textbf{12.81} & 11.79\\
\hline{}
Identity Mat. & 6.67 &  3.62\\
\end{tabular}
\\
\label{table:gcn_glove}
\end{wraptable}

In evaluating \#1, we compare 3 versions of Ego-OMG: one containing only the GCN stream; one containing only the CSN stream; one containing both the GCN and the CSN stream. We evaluate these configurations over both the seen and unseen test sets of EPIC Kitchens. As shown in Table \ref{table:gcn_csn_s1_s2}, CSN + GCN outperforms ablations. We also evaluate how performance for these configurations degrade across longer anticipation time $\tau_a$. Performance is shown in Table \ref{table:anticipation_time_vary}. We observe the performance of the GCN stream degrades gracefully with increasing $\tau_a$, outperforming the CSN stream over anticipation time $\tau_a = 5$.



Table \ref{table:anticipation_time_vary} shows our results. Note the joint model CSN + GCN outperforms either CSN or GCN alone over all anticipation times $\tau_a$, and the GCN stream outperforms the CSN stream when anticipating actions over the longest anticipation time $\tau_a = 5$ seconds. 

In evaluating \#2, we compare the use of an LSTM in Ego-OMG to two alternatives: 1) an average pooling of graph embedding features for nodes $S_k = \{s_1, s_2, ..., s_n\}$ for test video $V_k$, disregarding the sequential nature of the trajectory (``Mean Aggregation''), and 2) directly classifying the node embedding of the final state $s_n \in S_k$, ignoring the state history $s_{1:n-1}$ (``Terminal State Classification''). As shown in Table \ref{wrap-tab:1}, the sequential ordering of the LSTM outperforms the alternatives.

In evaluating \#3, we compare two versions of Ego-OMG: One where graph nodes $V$ are embedded through a GCN, and the other where nodes $V$ are left unaltered before the node sequence observed from the video clip is fed into the LSTM.

In evaluating \#4, we compare two versions of Ego-OMG:
One where the initialization of input matrix $X$ is set to features extracted from a pre-trained GloVe-600 model through methods discussed in Section \ref{sec:structured_graph_of_egocentric_activity}, and the other where $X$ is set to the identity matrix.


Table \ref{table:gcn_glove} illustrates results over joint ablations for \#3 and \#4. The use of graph convolutions in conjunction with GloVe embeddings outperforms ablations.



\section{Conclusion}
\label{sec:conclusion}
We have introduced Ego-OMG, a novel action anticipation method including representations for all of semantic temporal structure of activities, short term dynamics, and appearance of actions. Ego-OMG makes use of a novel graph representation capturing action structure at all three levels, whose nodes are embedded into a natural vector representation through use of a Graph Convolutional Network. This graph constitutes the core component of the first stream of Ego-OMG. Ego-OMG's second stream consists of CSN features. This second stream is easily swappable with other representations, making Ego-OMG complementary to many existing alternative action anticipation approaches. 



Ego-OMG attains state-of-the-art performance over the EPIC Kitchens Action Anticipation Challenge, and outperforms previous published architectures by large margins. 

\section*{Broader Impact}

Ego-OMG was designed with the intention of furthering the state of action anticipation methods on a path towards applications such as fluid human-robot interaction. Long-term, methods for better anticipation of action benefit those who would benefit from fluid interactions with robots - e.g., in easing manual cooperative tasks, such as in elder care. It would disadvantage those whose employment would be jeopardized by advances in robotics and automation. The consequences of the failure of Ego-OMG are application dependent. In this work, Ego-OMG is trained over EPIC Kitchens, but Ego-OMG is applicable to any egocentric activity data. EPIC Kitchens biases include 1) inclusion only of subjects of a socioeconomic status to have a kitchen, 2) all subjects are drawn from four cities in North America and Europe, and 3) the geographical distribution of subjects implies a bias in cuisine.











\bibliographystyle{unsrt}
\bibliography{neurips_2020}

\end{document}